\documentclass{article}

\usepackage[nonatbib, preprint]{neurips_2025}

\usepackage[utf8]{inputenc} 
\usepackage[T1]{fontenc}    
\usepackage{hyperref}       
\usepackage{url}            
\usepackage{booktabs}       
\usepackage{amsfonts}       
\usepackage{nicefrac}       
\usepackage{microtype}      
\usepackage{xcolor}    
\usepackage{siunitx}   
\usepackage{caption}   
\usepackage{makecell}

\usepackage{graphicx}
\usepackage[table]{xcolor} 
\usepackage{booktabs} 

\title{EgoSocial: Benchmarking Proactive Intervention Ability of Omnimodal LLMs via Egocentric Social Interaction Perception}

\author{%
  \makecell{Xijun Wang$^1$, Tanay Sharma$^2$, Achin Kulshrestha$^2$, Abhimitra Meka$^2$, Aveek Purohit$^2$, \\ Dinesh Manocha$^1$} \\
  $^1$University of Maryland, College Park\\
  $^2$Google\\
  \texttt{xijun@umd.edu} \\
}

\begin{document}

\maketitle

\begin{abstract}
As AR/VR technologies become integral to daily life, there's a growing need for AI that understands human social dynamics from an egocentric perspective. However, current LLMs often lack the social awareness to discern \textbf{when} to intervene as AI assistant. This leads to constant, socially unaware responses that may disrupt natural conversation and negatively impact user focus. 
To address these limitations, we introduce EgoSocial, a large-scale egocentric dataset with 13,500 social video-question pairs, specifically designed to benchmark intervention in social interaction perception. We also present an in-depth analysis of current omnimodal large multi-modal models (OLMMs) to assess their effectiveness in detecting diverse social contextual cues. Experiments show that OLMMs still struggle to detect the intervention timing (14.4\% for Gemini 2.5 Pro). We also propose EgoSoD (EgoSocial Detection), an end-to-end method for robustly discerning social dynamics. Informed by our OLMM analysis, EgoSoD integrates multimodal contextual cues (e.g., audio and visual cues) into a social thinking graph, dynamically modeling participants and interactions. Our method proactively detects intervention timing and social interactions, precisely determining when to intervene. From our experiments, our EgoSoD improves Phi-4 by 45.6\% and Gemini 2.5 Pro by 9.9\% on Intervention Timing performance, and improves Phi-4 by 20.4\% and Gemini 2.5 Pro by 6.9\% on overall Social Interaction performance. We will release the dataset and code soon.
\end{abstract}

\section{Introduction}
\begin{figure}[htb]
    \centering
    \includegraphics[width=1.\linewidth]{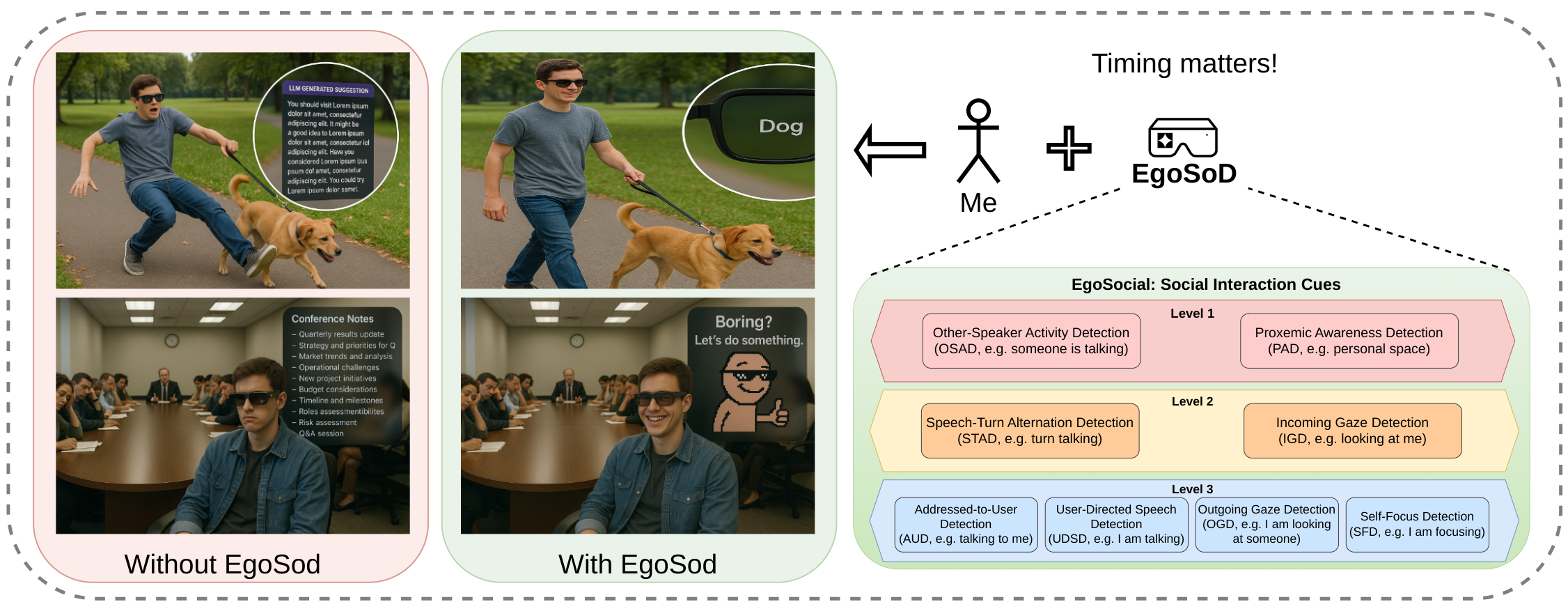}
    \vspace{-3mm}
    \caption{Task Overview. We define 8 social interaction cues from coarse environmental cues to fine-grained personal attentional state cues covering both audio and vision. Our EgoSoD helps LLMs make intelligent timing decisions for intervention.}
    \label{fig:task}
    \vspace{-6mm}
\end{figure}
The rapid advancement of Large Language Models (LLMs) \cite{zhang2024mm}, like ChatGPT~\cite{achiam2023gpt}, Gemini \cite{team2023gemini}, Claude~\cite{Anthropic2024Claude3}, DeepSeek~\cite{guo2025deepseek}, etc. has revolutionized our ability to understand and generate human-like text. As AR/VR technologies like headmounted displays and smart glasses become integral to our daily lives, these LLM-powered agents are increasingly integrated into various applications designed to enhance social interaction \cite{jiang2023social, xu2024can, yang2025socialmind}. Such applications are already mediating human-to-human communication through real-time translation, meeting summaries, or subtle suggested replies on smart glasses in social scenarios \cite{yang2025socialmind}.

However, most current interactions (i.e. ask and respond) are user-initiated. While effective, a key challenge emerges when these LLMs operate in an "always-on" mode, needing to proactively respond without explicit prompts. This often leads to unintended responses by LLMs at inappropriate times. Human social interactions are incredibly complex, relying on a dynamic interplay of verbal and non-verbal cues, rapid turn-taking, and an unspoken understanding of conversational rhythm. While OLMMs have made impressive strides in determining  \textit{what} to respond, existing research rarely addresses \textit{when} an OLMM should proactively intervene. This precise timing, often subtle yet critical, remains an underexplored dimension in socially intelligent systems. Accurately detecting social interactions and understanding relevant cues is essential for OLMM agents to decide when to process context and respond without being intrusive or disrupting the natural flow of conversation.

Current methods for proactive social assistance face significant hurdles. Continuous or frequent suggestions \cite{yang2025socialmind}—whether visual or audio—can disrupt natural conversation, negatively impacting user engagement and cognitive flow. Another issue is the reliance on numerous subsystems within traditional cascaded architectures, which often leads to accumulated errors and compromises overall system performance \cite{pan2022quantifying}.Crucially, the absence of standardized egocentric dataset benchmarks makes it challenging to quantitatively evaluate these proactive assistance systems in real social interaction scenarios. Many existing social interaction datasets are exocentric (third-person view) \cite{muller2022multimediate, raman2022conflab, lai2022werewolf, jahangard2024jrdbsocialmultifacetedroboticdataset, zhan2023socialdial}, rendering them unsuitable for emerging AR/VR or wearable devices in this domain. Moreover, even available egocentric datasets~\cite{aghaei2016whom} often lack comprehensive multimodal annotations, offering only single modalities (e.g., photo-streams, text, or audio). If they do include audio and video, they typically lack the specific cue annotations crucial for understanding complex social dynamics. 

To address these issues, in this paper, we first analyze the social interaction survey~\cite{barua2021detecting} and the effectiveness of current OLMMs, which unify text, vision, and audio in one model. Based on this analysis, we define eight social interaction cues and introduce the EgoSocial dataset. This comprehensive resource comprises 13,500 video-question pairs within 10-second segments, aiming to enable rigorous benchmarking for intervention and social interaction detection.


Based on the OLMM analysis, we also present EgoSoD (EgoSocial Detection), a novel proactive social interaction detection method. EgoSoD distinguishes itself from conventional systems by aiming for minimal-disturbance intervention, assisting users only when truly necessary to reduce distraction and maintain natural conversation flow. This is achieved through an integrated social graph-based reasoning module, which dynamically incorporates contextual elements, including interaction scenarios, participants' spatial relationships (e.g., proximity), visual cues (e.g., eye-contact), and audio cues (e.g., speech activity, speaker identification). We demonstrate the effectiveness of our EgoSoD by identifying and incorporating the right cues. Our dataset and approach will expand this domain, shifting research focus toward when LLM models should respond within egocentric video contexts.

Our contributions can be summarized as follows:
\begin{itemize}
\item \textbf{Concept Definition}: We're the first, to our knowledge, to clearly define the social cues for detecting social interventions/interactions and to enable their quantitative evaluation.
\item \textbf{EgoSocial Dataset}: We introduce EgoSocial, a large-scale evaluation benchmark comprising 13,500 manually annotated video-question pairs. 
\item \textbf{EgoSoD Framework}: We propose EgoSoD, a novel thinking-graph-based method that captures rich contextual features to guide OLMMs in making intelligent timing decisions for intervention.
\end{itemize}

\section{Related Works}
\subsection{Social Interaction Related Benchmarks}
Research into understanding human social behavior via computer vision and machine learning heavily depends on benchmark datasets. However, these datasets vary significantly in their characteristics and utility for robust social interaction detection. While some existing datasets focus on social group detection or pattern characterization, they often lack sufficient interaction sequences from multiple viewpoints or miss the diverse cue annotations necessary for perceiving nuanced interactions in natural settings. Although benchmarks for socially aware dialogue systems have been explored \cite{zhan2023socialdial}, they primarily deal with dialogue data in text modality. Crucially, there is a scarcity of standardized, publicly available multimodal egocentric datasets specifically tailored for this task. While Ego4D offers valuable egocentric conversational videos, it requires a specialized sampling and annotation strategy to create a balanced dataset suitable for robust social interaction detection and analysis. This strategy must ensure a wider distribution of social cues across various durations, critically including samples where no conversation or interaction occurs to provide essential negative examples. The ultimate aim is to provide appropriate data that enables current omnimodal LMMs to learn precisely when to respond in social contexts.

\begin{table}[h!]
\centering
\small
\label{tab:datasets}
\begin{tabular}{lcccc} 
\toprule
\textbf{Dataset Name} & \textbf{View} & \textbf{Modalities} & \textbf{Social Cues} & \textbf{Intervention Det.} \\
\midrule
MPIIGroupInteraction \cite{muller2022multimediate} & Exocentric & Video, Audio & No & No \\
JRDB-Social \cite{jahangard2024jrdbsocialmultifacetedroboticdataset} & Exocentric & Video, Audio, Multi-sensor & No  & No \\
Werewolf Among Us \cite{lai2022werewolf} & Exocentric & Video, Audio, Text & No  & No\\
ConfLab \cite{raman2022conflab} & Exocentric & Video, Audio, Multi-sensor & No  & No\\
SocialDial \cite{zhan2023socialdial} & Exocentric & Text & No  & No\\
EgoSocialStyle \cite{aghaei2018towards} & Egocentric & Image & No  & No\\
Ego4D \cite{grauman2022ego4d} & Egocentric & Video, Audio, Text & Partial  & No\\
\textbf{EgoSocial (Ours)} & Egocentric & Video, Audio, Text & Yes  & Yes \\
\bottomrule
\end{tabular}
\vspace{1mm}
\caption{Overview of Datasets. We are first to clearly define the social cues for social intervention/interaction detection and enable quantitative evaluation.}
\vspace{-4mm}
\end{table}

\subsection{Social Interaction Detection}

Egocentric social interaction detection, which involves identifying when a camera wearer is conversing or interacting with others from a first-person perspective, has been a significant area of focus in computational social science and computer vision. Early approaches typically relied on traditional computer vision and machine learning methods \cite{aghaei2016whom}, often leveraging sociological concepts like F-formation by analyzing individuals' distance and orientation relative to the camera wearer \cite{aghaei2016whom}. Temporal modeling, frequently employing LSTMs or GRUs, was also common to capture the evolution of these social signals over time \cite{felicioni2021interaction}.

Subsequent research expanded the range of visual cues to include head pose, attention patterns, and facial expressions \cite{felicioni2021interaction, yang2025socialmind}. These cues were sometimes integrated into frameworks like Graph Convolutional Networks (GCNs) to model relationships between people \cite{felicioni2021interaction}. Beyond visual data, some studies explored approaches using only audio modality \cite{dieperink2023investigating}, while others investigated multimodal fusion techniques combining both visual and audio information for tasks such as "Talking to me" detection \cite{nadar2024perspective}. Further research also focused on detecting more subtle cues or micro-actions within dyadic interactions, emphasizing the value of paired egocentric perspectives \cite{yonetani2016recognizing}.

While these methods have demonstrated capabilities in detecting or classifying social interactions \cite{felicioni2021interaction, nadar2024perspective}, they predominantly rely on explicit feature engineering and specialized models tailored for specific signal processing and temporal/relational modeling \cite{aghaei2016whom}. Crucially, these approaches do not directly evaluate large language models (LLMs) or vision-language models (VLMs) for the primary task of social interaction detection from these foundational signals. Although recent work explores using VLMs for extracting contextual cues in tasks like gaze following \cite{gupta2024exploring}, or employs LLMs for reasoning over pre-extracted social cues and context within social assistive systems \cite{yang2025socialmind}, they generally do not investigate LLMs' inherent capability to detect the interaction event itself or analyze the direct importance of different social signals within the LLM framework for detection in challenging scenarios. Existing LLM applications often rely on direct prompting without a detailed examination of their foundational capabilities in this domain.

\subsection{Omnimodal LMMs}

The landscape of Large Multimodal Models (LMMs) has rapidly evolved, transitioning from text-only to encompassing diverse modalities, culminating in omnimodal LMMs like GPT-4o and Google's Gemini \cite{team2023gemini}. While early models focused on two modalities (e.g., LLaVA \cite{lin2023video}, Whisper \cite{radford2023robust}), the trend now favors unified architectures that seamlessly integrate multiple inputs, driven by the pursuit of comprehensive reasoning. Smaller models like Microsoft's Phi-4-multimodal also demonstrate robust multimodal capabilities on edge devices. However, despite their "academic intelligence," current LMMs often struggle with social intelligence \cite{xu2024academically}. As LMMs become integrated into agentic systems, it's crucial for them to move beyond constant responses and demonstrate a nuanced understanding of social norms, recognizing when silence is appropriate, thereby fostering more intuitive and less intrusive human-AI collaboration. 
\section{Benchmark: EgoSocial}
\subsection{Data curation}
\textbf{Data processing.}
We constructed our dataset based on videos from the Ego4D~\cite{grauman2022ego4d} dataset. These videos capture diverse real-world egocentric multi-user scenarios, including games, eating/drinking etc. Each original Ego4D clip is 5 minutes long. To provide optimal context for OLMM models to make real-time decisions and facilitate standardised analysis, we segmented these clips into 10-second chunks. This duration offers sufficient context for models to make informed decisions without becoming computationally overwhelming. We downsampled the video frequency to 1 frame per second (1fps) to align with real-time processing requirements. This resulted in a refined dataset comprising 1500 video segments. \\ 
\begin{figure}[h]
    \centering
    \includegraphics[width=1.\linewidth]{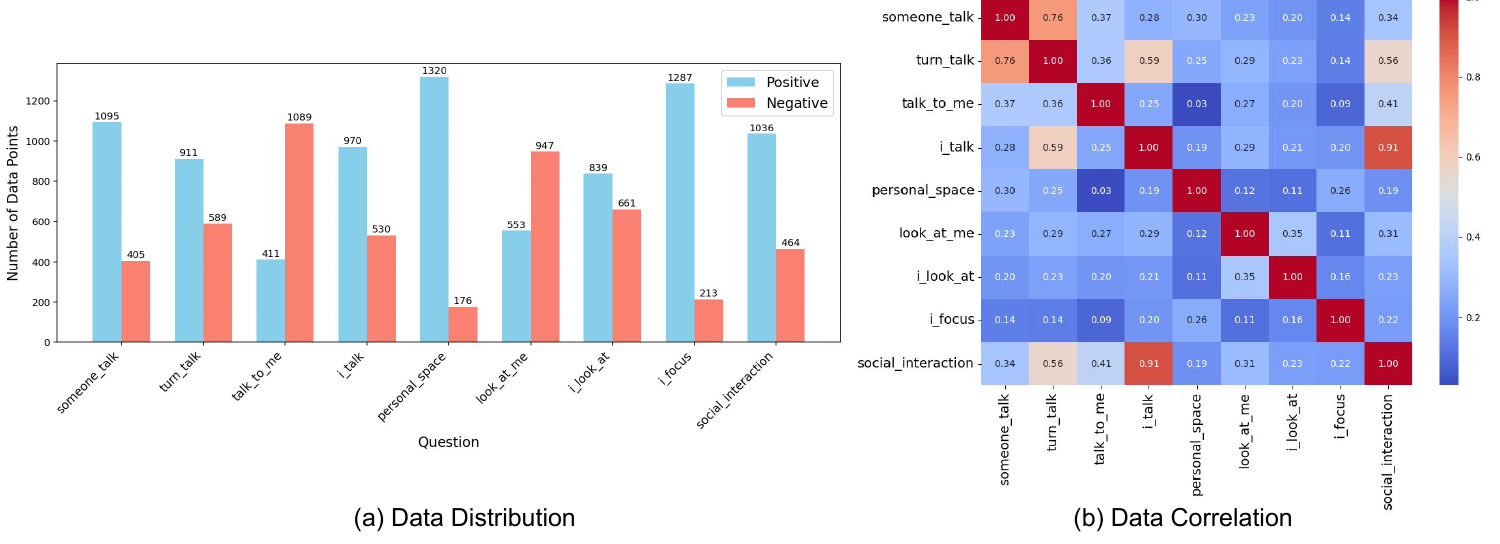}
    \caption{Data distribution and correlation on our EgoSocial dataset.}
    \label{fig:data_distribution}
    \vspace{-4mm}
\end{figure}

\textbf{Data annotation.}
We detail our annotation process for establishing ground truth and identifying 8 social cues per video segment, combining three existing Ego4D annotations with five new manually labeled ones. Our primary objective defines social interaction as an active, undisturbed conversation where "me" (the glasses wearer) is speaking or being spoken to directly; other scenarios, including challenging group settings without direct wearer participation, are considered negative samples. Figure~\ref{fig:data_distribution} shows the distribution of cues and ground truth. This process applied to 1500 video segments, generated 13,500 video-question pairs, aimed at analyzing current OLMM models' understanding of complex social dynamics. More comprehensive annotation details and guidelines are in the supplementary material.

\subsection{Cue Correlation Analysis}
In this section, we analyzed the interrelationships and individual significance of our social cues, as shown in Figure~\ref{fig:data_distribution}. We found a moderate positive correlation (0.35) between "Am I looking at someone?" and "Is someone looking at me?", suggesting a mutual gaze dynamic, though not consistently strong. Interestingly, "Are people in personal space?" showed weaker correlations with most cues, its highest being a moderate 0.30 with "Is someone talking?". This highlights that proximity doesn't always equate to active social interaction. Crucially, every individual cue exhibited a positive relationship with our social interaction detection task, leading us to retain all of them as each contributes valuable information.

\section{Method: EgoSoD}
\label{method}

In this section, we'll explain EgoSoD's core framework, detailing how it determines the optimal moment to intervene. We'll describe the key detections EgoSoD performs automatically based on our identified social cues. We'll also discuss how these detections are utilized in a hierarchical manner.


\begin{figure}[t]
    \centering
    \includegraphics[width=1.\linewidth]{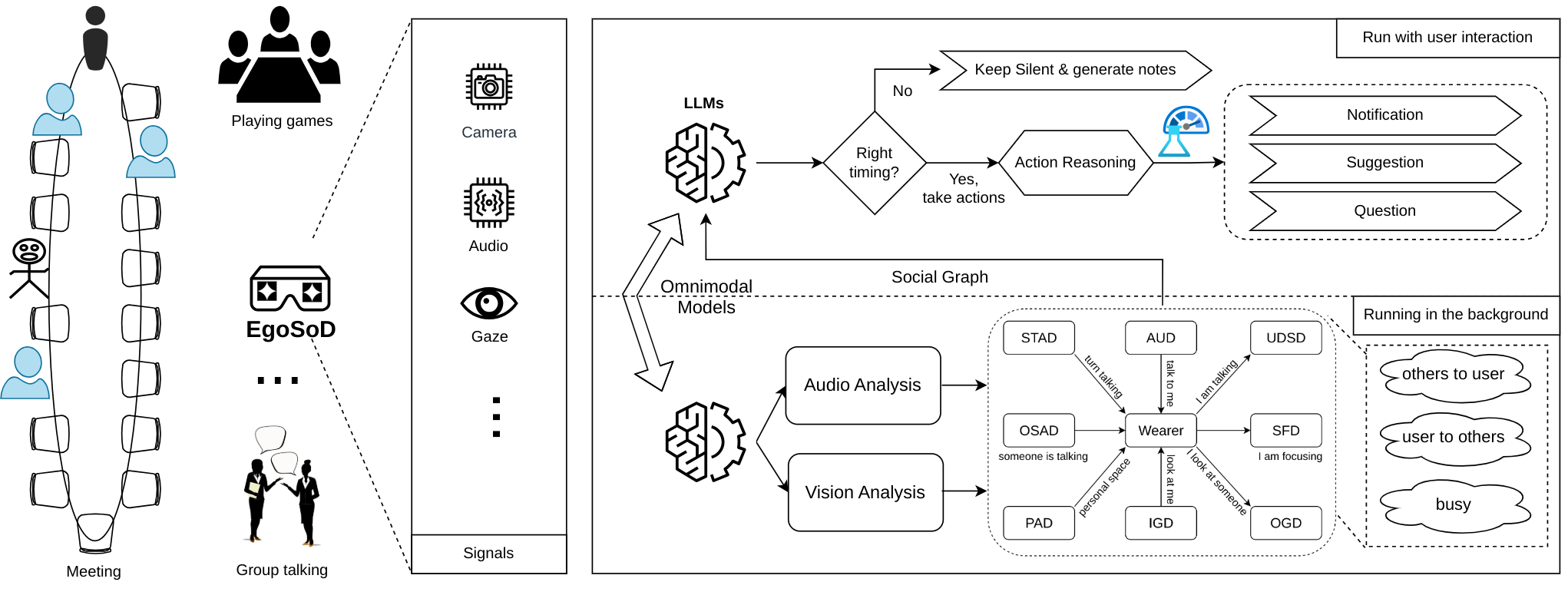}
    \vspace{-4mm}
    \caption{EgoSoD Overview. We propose a thinking-graph-based method that captures rich contextual features to guide OLLMs in making intelligent timing decisions for intervention.}
    \label{fig:egothink}
    \vspace{-4mm}
\end{figure}
\subsection{Audio Analysis}

To provide our system with robust conversational awareness, we've designed a four-level audio-processing pipeline, powered by OLMMs. This pipeline efficiently transforms raw acoustic input into meaningful social cues, progressing from general scene monitoring to fine-grained, directed-speech understanding, all integrated within the OLMM's logic:

\textbf{Other-Speaker Activity Detection (OSAD)}
It first detects any voice activity. This initial check prevents unnecessary processing when the environment is silent.

\textbf{Speech-Turn Alternation Detection (STAD)}
It identifies when one speaker stops and another begins. Spotting these genuine back-and-forth patterns, rather than just monologues, signals an active dialogue. This also tells us the wearer might need to pay attention, as their turn could be next.

\textbf{Addressed-to-User Detection (AUD)}
For every turn by a non-wearer, it determines if the speech is directed at the wearer. This helps the system focus on relevant social signals and ignore background noise. Positive AUD decisions can trigger alerts, notifying the user of incoming attention. This precision is vital for accurate, personalized support and avoiding false alarms.

\textbf{User-Directed Speech Detection (UDSD)}
Symmetrically, it monitors the wearer's own speech to see if they're addressing someone else. This ensures contextual assistance is only offered after the user finishes speaking, for example, after answering a question. UDSD events can also prepare higher-level applications like task assistance or note-taking.

While acoustic cues drive this hierarchy, OLMMs are superior at understanding language. Therefore, it converts each utterance into text, which serves as a stable "anchor" for its own interpretation. These transcripts not only provide linguistic features for AUD/UDSD processing but also allow the OLMM to understand references, fill in missing words, and integrate information from multiple sources.

By leveraging OLMMs to perform OSAD, STAD, AUD, and UDSD, and anchoring their internal processing in text, our pipeline equips EgoSoD with a deep understanding of conversations, both structurally and semantically. This hierarchical design significantly improves our grasp of audio scenes, minimizes unnecessary interventions, and ensures the OLMM-driven modules respond only when truly helpful.

\subsection{Visual Analysis}

Our Vision Processing module is organized as a four-stage hierarchy that interprets visual input. This progressive design moves from broad environmental cues to precise attentional states, ensuring early filtering of irrelevant scenes and reliable social interaction understanding.

\textbf{Proxemic Awareness Detection (PAD)}
It first quickly estimates if any person is within the wearer’s personal space (around 1.2 meters). Scenes without this proximity are immediately filtered, saving computation by focusing only on potential interactions.

\textbf{Incoming Gaze Detection (IGD)}
Once proximity is confirmed, it evaluates whether nearby individuals are looking directly at the wearer. Direct gaze strongly signals potential verbal or gestural interaction. Detecting this allows the system to predict potential social interactions and prepare relevant information for the wearer.

\textbf{Outgoing Gaze Detection (OGD)}
It then shifts focus to the wearer’s own gaze, determining if they are looking at another person and if this attention is sustained. Identifying this reciprocal or directed gaze clarifies intentional engagement from mere coincidence, refining the system's understanding of who the likely conversational partner is. 

\textbf{Self-Focus Detection (SFD)}
Finally, it identifies if the wearer is internally focused (e.g., using a device or handling objects). This is crucial: recognizing such self-focus prevents disruptions, ensuring assistance waits until the wearer is ready. By combining this detailed attentional state with spatial and gaze cues, the system gains a complete visual understanding, allowing for truly appropriate interventions.

In summary, our visual pipeline uses Proxemic Awareness to first filter out distant scenes. Then, it layers Incoming Gaze to spot social invites, Outgoing Gaze to confirm shared attention, and Self-Focus Recognition to check if the user is available. This tiered approach efficiently moves from broad spatial cues to detailed attention states, ensuring the system only steps in at the right moment, subtly and effectively.

\subsection{Social-Thinking Graph for Multimodal Interaction Reasoning}

To enable social reasoning, we structure the eight cues discussed previously into a \textit{Social-Thinking Graph}. Audio is processed by OSAD, while egocentric video frames are analyzed by PAD. OSAD and PAD act as initial filters: if no speech is present or no one is close, the corresponding processing branch stops for that moment, saving computation.

When these initial filters are open, the system looks for early signs of engagement. STAD confirms genuine back-and-forth turns in the audio, while IGD checks if any nearby person is looking directly at the wearer. A positive detection from either of these cues upgrades the context to a possible interaction, unlocking more detailed reasoning. Next, four detectors identify speaker roles and the wearer’s availability. AUD spots non-wearers talking to the wearer. Symmetrically, UDSD recognizes if the wearer is speaking to someone else. In the visual stream, OGD determines if the wearer maintains eye contact, while SFD infers inward absorption from cues like hand-object interactions. A positive SFD acts as a global veto, suppressing prompts when the wearer is cognitively busy.

Then combined into three belief variables:
\[
\mathrm{Others\to User}= \mathrm{PAD} \land \mathrm{IGD} \land \mathrm{AUD},
\]
\[
\mathrm{User\to Others}= \mathrm{STAD} \land \mathrm{UDSD} \land \mathrm{OGD},
\]
\[
\mathrm{UserBusy}= \mathrm{SFD}.
\]
The model uses these three variables to decide if the user is engaged in a social interaction. This \textit{Social-Thinking Graph} provides an interpretable mechanism for detecting social interactions.

\section{Experiments}
\subsection{Experimental Setup}
Omni-modal large multi-modal models (OLMMs) unify text, vision, audio—and in the newest releases, 3D and code—inside a single transformer backbone. Because social interaction detection depends on jointly interpreting speech content and prosody, eye-contact and body pose, spatial proximity, and even visual scene context, a unified model can reason across modalities without the cascading errors that plague specialised pipelines. We apply Phi-4-multimodal-instruct 5.6B ~\cite{abouelenin2025phi}, Gemini 1.5 Pro~\cite{team2024gemini}, Gemini 2.0 Flash~\cite{google_gemini2}, and Gemini 2.5 Pro~\cite{google_deepmind_gemini_2_5}. These four OLMMs span \textit{open vs.\ proprietary}, \textit{small vs.\ large}, and \textit{speed vs.\ reasoning-oriented} trade-offs. For Intervention Timing, we use the accuracy as metric. For overall Social Interaction, we use Macro F1 score.

\subsection{Results}
\begin{table}[tbp] 
\centering
\small
\begin{tabular}{lcccc}
\toprule
\textbf{Models} & {\textbf{Video}} & {\textbf{Audio+Video}} & {\textbf{Audio+Video+Text}} & {\textbf{Audio+Video+Text w.Conv}} \\
\midrule
\multicolumn{5}{c}{\textit{Intervention Timing}} \\
\cmidrule(lr){1-5} 
Phi-4        & 11.85 & 12.50 & 31.47 & 31.47 \\
Gemini 1.5 Pro & 14.66 & 13.36 & 13.79 & 13.36 \\
Gemini 2 Flash   & 16.81 & 17.67 & 21.98 & 20.69 \\
Gemini 2.5 Pro & 14.66 & 14.44 & 19.18 & 20.69 \\
\midrule
\multicolumn{5}{c}{\textit{Overall Social Interaction: Macro F1}} \\
\cmidrule(lr){1-5}
Phi-4        & 50.56 & 51.65 & 63.45 & 64.61 \\
Gemini 1.5 Pro & 51.84 & 53.16 & 53.48 & 53.12 \\
Gemini 2 Flash   & 52.91 & 56.08 & 58.75 & 58.01 \\
Gemini 2.5 Pro & 52.46 & 53.97 & 56.19 & 57.75 \\
\bottomrule
\end{tabular}
\vspace{1mm}
\caption{Social Interaction Detection Performance with different modalities on EgoSocial. We observed that adding raw audio modality to video inputs generally improved the overall social interaction performance, while not all models benefit on intervention performance. Audio to text benefit all models. (w.Conv means in conversation format with speaker labels.)}
\label{tab:modality_improved} 
\vspace{-7mm}
\end{table}
We first explore the importance of different modalities in OLMM analysis, followed by an assessment of OLMMs' inherent ability to detect various social cues. We then analyze how individual social cues correlate with the overall social interaction outcome, achieved by conditioning the model to focus on each cue separately. Finally, we present the performance of our proposed EgoSoD framework and compare its effectiveness against established baseline methods.

\subsubsection{Impact of modality on social interaction detection}
We conducted experiments on various OLMMs to detect social interaction using different modalities, as detailed in Table \ref{tab:modality_improved}. Our evaluation focused on two key metrics: the overall ability to detect Intervention Timing (Accuracy) and
Social Interaction (Macro F1).

Initially, we observed that adding an audio modality to video inputs generally improved the overall social interaction performance, while not all models are benefit it on intervention performance. 
%
%
To further investigate, we converted the audio to text (transcriptions) and included it as a separate modality. This yielded a significantly higher performance gain across all models compared to using raw audio alone. 
This is a crucial finding because, despite speech being fundamental to social interaction, these LLMs currently struggle to effectively decipher social cues directly from audio signals, performing much better when speech is processed as text.
%
%
Across all models, we consistently observed a much higher Macro F1 accuracy for overall social interaction detection compared to the intervention timing detection. 
This wide gap indicates that current OLLMs, while capable of detecting that a social interaction is occurring, still struggle to pinpoint the precise moment for intervention.

\subsubsection{Social Cues Detection Results}
Our analysis of social cue detection by OLMMs, summarized in Table \ref{tab:signals_improved}, reveals varying strengths and persistent challenges. Models generally perform well in detecting fundamental conversational structures like Other-Speaker Activity Detection (OSAD) and Speech-Turn Alternation Detection (STAD). 

However, LLMs demonstrate significant difficulty with more nuanced cues. Addressed-to-User Detection (AUD), discerning if speech is directed at the wearer, shows consistently low performance (e.g., Gemini 1.5 Pro at 28.65\%), highlighting a struggle with subtle speaker-listener dynamics. Similarly, Outgoing Gaze Detection (OGD)—the wearer looking at someone else—is poorly recognized, with overall accuracy often below 50\% (e.g., Gemini 2.5 Pro at 47.15\%), indicating a general inability to accurately interpret the wearer's attention towards others.


Finally, model-specific performance insights are noteworthy. Phi4, despite its smaller size, consistently performs competitively, often leading in crucial cues like Proxemic Awareness Detection (PAD) (78.47\%), which proved challenging for Gemini 2.5 Pro (38.16\%). This suggests that smaller, highly optimized architectures could be particularly effective for egocentric distance estimation.

\begin{table}[htbp] 
\centering
\small
\resizebox{1\textwidth}{!}{
\begin{tabular}{lcccccccc}
\toprule
\textbf{Model} & {\textbf{OSAD ($\uparrow$)}} & {\textbf{STAD($\uparrow$)}} & {\textbf{AUD($\uparrow$)}}& {\textbf{UDSD($\uparrow$)}}& {\textbf{PAD($\uparrow$)}}& {\textbf{IGD($\uparrow$)}}& {\textbf{OGD($\uparrow$)}}& {\textbf{SFD($\uparrow$)}} \\
\midrule
\multicolumn{9}{c}{\textit{Positive Data points Performance}} \\
\cmidrule(lr){1-9} 
Phi-4        & 77.17 & 97.15 & 77.86 & 88.04 & 98.79 & 86.08 & 99.76 & 84.69 \\
Gemini 1.5 Pro & 85.57 & 75.52 & 96.59 & 96.91 & 93.26 & 84.62 & 99.88 & 51.90 \\
Gemini 2 Flash   & 86.03 & 87.60 & 96.36 & 93.92 & 93.71 & 47.56 & 99.76 & 65.35 \\
Gemini 2.5 Pro & 90.77 & 81.45 & 92.70 & 86.08 & 38.48 & 46.47 & 98.93 & 94.17 \\
\midrule
\multicolumn{9}{c}{\textit{Negative Data points Performance}} \\
\cmidrule(lr){1-9}
Phi-4        & 72.59 & 39.39 & 57.39 & 48.87 & 47.78 & 33.58 &  6.66 & 23.94 \\
Gemini 1.5 Pro & 67.16 & 77.26 & 28.65 & 38.49 & 76.67 & 41.50 &  5.90 & 43.66 \\
Gemini 2 Flash   & 72.84 & 57.89 & 29.66 & 40.94 & 61.67 & 81.20 &  6.35 & 58.69 \\
Gemini 2.5 Pro & 73.83 & 84.08 & 37.28 & 55.66 & 71.59 & 94.93 & 11.65 & 29.11 \\
\midrule
\multicolumn{9}{c}{\textit{Overall Macro-F1 Performance}} \\
\cmidrule(lr){1-9}
Phi-4        & 72.18 & 68.50 & 61.40 & 69.38 & 78.47 & 52.41 & 42.73 & 54.00 \\
Gemini 1.5 Pro & 75.85 & 75.61 & 47.10 & 68.74 & 81.38 & 57.29 & 42.01 & 42.25 \\
Gemini 2 Flash   & 78.46 & 73.47 & 47.81 & 68.45 & 76.79 & 64.79 & 42.43 & 53.90 \\
Gemini 2.5 Pro & 82.43 & 81.98 & 52.45 & 71.66 & 38.16 & 71.92 & 47.15 & 63.45 \\
\bottomrule
\end{tabular}
}
\vspace{1mm}
\caption{Social Cues Detection on EgoSocial. We show results on positive data accuracy, negative data accuracy, and the overall Macro-F1 scores across eight social cue detection tasks. }
\label{tab:signals_improved} 
\vspace{-6mm}
\end{table}


\subsubsection{Single Cue-guided Social Interaction Detection Results}
To understand how individual social cues influence both overall social interaction and intervention timing, we conducted experiments where LLMs were explicitly guided by one specific signal to reason about the final outcome, as shown in Table \ref{tab:si_signal_improved}.

A clear and consistent improvement is observed across all models for both overall Macro F1 and Timing when guided by a single cue, compared to their unguided baseline performance. This indicates that providing LLMs with focused contextual information significantly enhances their ability to detect social interactions and, crucially, to discern the optimal moment for intervention.

For instance, Gemini 1.5 Pro achieves its highest Macro F1 score of 74.52\% for overall social interaction and a robust 55.39\% Timing Accuracy when specifically guided by Addressed-to-User Detection (AUD). This highlights the strong positive correlation between being directly addressed and both overall interaction detection and the precision of intervention timing for this model. This also addresses a critical weakness observed at baseline: providing LLMs with explicit guidance on individual social cue substantially improves their ability to correctly identify when social interaction is absent (i.e., negative detection), leading to more balanced performance in both metrics.

While guidance generally helps improve timing accuracy, the degree of improvement varies. Even within the same model family, a cue like AUD might lead to a strong gain for Gemini 1.5 Pro's timing (55.39\%), while other cues show less pronounced effects for other models (e.g., Gemini 2.5 Pro's timing performance only reaching 40.52\% with STAD). This variability underscores that while individual cues are valuable, their utility in guiding LLM reasoning for precise intervention timing is nuanced and model-dependent.

\begin{table}[tbp] 
\centering
\small 
\resizebox{1\textwidth}{!}{
\begin{tabular}{lccccccccc}
\toprule
\textbf{Model} & {\textbf{Baseline}} & {\textbf{OSAD }} & {\textbf{STAD}} & {\textbf{AUD}}& {\textbf{UDSD}}& {\textbf{PAD}}& {\textbf{IGD}}& {\textbf{OGD}}& {\textbf{SFD}} \\
\midrule
\multicolumn{10}{c}{\textit{Intervention Timing}} \\
\cmidrule(lr){1-10} 
Phi-4      & 12.50   & 32.11 & 44.18 & 33.62 & 34.91 & 25.22 & 30.17 & 29.53 & 29.09 \\
Gemini 1.5 Pro  & 13.36 & 32.33 & 29.96 & 55.39 & 17.03 & 14.87 & 17.03 & 13.79 & 13.79 \\
Gemini 2 Flash  & 17.67  & 31.25 & 35.78 & 42.03 & 36.85 & 20.04 & 34.91 & 24.35 & 24.78 \\
Gemini 2.5 Pro & 14.44  & 25.65 & 40.52 & 35.34 & 21.98 & 21.12 & 24.14 & 20.47 & 21.55 \\
\midrule
\multicolumn{10}{c}{\textit{Overall Social Interaction: Macro F1}} \\
\cmidrule(lr){1-10}
Phi-4      & 51.65   & 63.51 & 68.25 & 64.25 & 65.32 & 60.60 & 63.26 & 62.45 & 62.97 \\
Gemini 1.5 Pro  & 53.16 & 65.51 & 63.13 & 74.52 & 55.67 & 53.86 & 55.63 & 53.39 & 53.34 \\
Gemini 2 Flash  & 56.08  & 64.18 & 66.61 & 68.30 & 66.29 & 55.54 & 65.38 & 59.47 & 60.25 \\
Gemini 2.5 Pro & 53.97 & 60.95 & 66.76 & 66.35 & 59.39 & 58.18 & 60.61 & 58.32 & 59.34 \\
\bottomrule
\end{tabular}
}
\vspace{1mm}
\caption{Single Cue-guided Social Interaction Detection Performance on EgoSocial. A clear and consistent improvement is observed across all models for both overall Macro F1 and Timing when guided by a single cue, compared to their unguided baseline performance.}
\label{tab:si_signal_improved} 
\vspace{-8mm}
\end{table}

\begin{table}[htbp]
\centering
\small
\begin{tabular}{lcccc}
\toprule
\textbf{Metrics / Models} & \textbf{Phi-4} & \textbf{Gemini 1.5 Pro} & \textbf{Gemini 2 Flash} & \textbf{Gemini 2.5 Pro} \\
\midrule
\multicolumn{5}{c}{\textit{Intervention Timing}} \\
\cmidrule(lr){1-5}
Baseline & 12.50 & 13.36 & 17.67 & 14.44 \\
EgoSoD (Ours) & 58.41 & 20.69 & 38.79 & 24.35 \\
\midrule
\multicolumn{5}{c}{\textit{Overall Social Interaction: Macro F1}} \\
\cmidrule(lr){1-5}
Baseline & 51.65 & 53.16 & 56.08 & 53.97 \\
EgoSoD (Ours)  & 72.08 & 58.64 & 67.83 & 60.88 \\
\bottomrule
\end{tabular}
\vspace{1mm}
\caption{Social Interaction Detection Performance with Graph-based Methods on EgoSocial (Transposed). Our method boosts Timing accuracy by 45.6\% on Phi-4 and 9.9\% on Gemini 2.5 Pro. Overall Macro F1 also improves by 20.4\% on Phi-4 and 6.9\% on Gemini 2.5 Pro.}
\vspace{1mm}
\label{tab:graph_improved_transposed}
\vspace{-6mm}
\end{table}

\subsubsection{Social Graph Results}
Our findings, based on the method described in Section \ref{method} and summarized in \ref{tab:graph_improved_transposed}, reveal several key insights into social interaction detection. We observed consistent gains in the "Timing" metric across all models when comparing our approach to the baseline. This is particularly noteworthy given that predicting when a model should interact with a user remains a significant challenge. The improvements suggest that combining various signals within our proposed framework effectively enhances the model's ability to tackle this complex task.

Furthermore, among the models evaluated in this final graph-based framework, Phi4 emerged as the top performer, demonstrating the best capability in integrating and interpreting the combined cues. This highlights its potential for making more accurate sense of complex social interaction data. While our method achieved a significant increase over baseline numbers and even single-cue guided performance, the absolute timing accuracy, despite the gains, still leaves room for improvement. This indicates a continuing need for further research and refinement in this area to achieve truly robust and seamless social interaction.

\section{Conclusion, Limitations and Future Work}
We introduced a novel dataset for social intervention/interaction detection, annotated with fine-grained social cues. Our analysis revealed that even advanced LLMs struggle with this task, particularly in detecting the correct timing to respond. To address this, we developed an extensible method that improved performance by effectively combining various signals. We conducted an extensive analysis of LLM behavior in this domain, noting that providing context hierarchically consistently enhanced their ability to understand cues and identify non-interactive moments.
The limitation is we don't have enough resource to test more Omnimodal models, and we will continue adding more models. In the future, we plan to go a further step to do 1 second level social intervention and interaction detection.

\appendix
\newpage

\section{Experiments Settings}
\paragraph{Intervention Timing Metric (ITM):} When we want to intervene users, users should be free or not involved in social interaction. So the precision for negative is the most important metric for us to decide if the user can be intervened or not.
\begin{equation}
    \text{ITM} = \frac{TP_{neg}}{TP_{neg} + FN_{neg}},
\end{equation}
where $TP_{neg}$ stands for True Positives for class negative. These are the instances that are actually of class negative and were correctly predicted as class negative. $FN_{neg}$ stands for False Negatives for class negative. These are the instances that are actually of class negative but were incorrectly predicted as belonging to a different class.

\paragraph{Social Interaction Metric (SIM):}
\begin{equation}
    \text{SIM} = \frac{1}{N} \sum_{i=1}^{N} \left( 2 \times \frac{\frac{TP_i}{TP_i + FP_i} \times \frac{TP_i}{TP_i + FN_i}}{\frac{TP_i}{TP_i + FP_i} + \frac{TP_i}{TP_i + FN_i}} \right),
\end{equation}
where $TP_i$ stands for True Positives for class i. These are the instances that are actually of class i and were correctly predicted as class i. $FP_i$ stands for False Positives for class i. These are the instances that are not of class i but were incorrectly predicted as class i. $FN_i$ stands for False Negatives for class i. These are the instances that are actually of class i but were incorrectly predicted as belonging to a different class.

\paragraph{Resources:}
All experiments have been finished with computing node with CPU: AMD Ryzen Threadripper PRO 3945WX 12-Cores; Memory: 130 GiB; GPU: $2 \times$ NVIDIA RTX 6000 Ada Generation, 49,140 MiB.

\section{Ablation Studies}
We first conduct ablation studies on the contributions of different components, and then explore different graph format and thinking paradigm, finally we test on how much visual information needed on this task.

\subsection{Different Components in EgoSoD}
To understand the impact of individual components in EgoSoD, we performed ablation studies focusing on the audio analysis (APG: Audio Processing Graph) and visual analysis (VPG: Video Processing Graph). The results, presented in Table~\ref{tab:comp}, indicate that both APG and VPG positively contribute to the overall performance. Notably, the audio analysis (APG) provides a stronger contribution.
\begin{table}[!h] 
\centering
\small
\begin{tabular}{lcccc}
\toprule
\textbf{Models} & {\textbf{Baseline}} & {\textbf{Baseline + APG}} & {\textbf{Baseline + VPG}} & {\textbf{Baseline + APG + VPG}} \\
\midrule
\multicolumn{5}{c}{\textit{Intervention Timing}} \\
\cmidrule(lr){1-5} 
Phi-4        & 12.50 & 46.12 & 24.57 & 58.41 \\
Gemini 1.5 Pro  & 13.36 & 15.52 & 14.66 & 20.69 \\
Gemini 2 Flash   & 17.67 & 29.96 & 19.61 & 38.79 \\
Gemini 2.5 Pro   & 14.44 & 24.35 & 15.30 &  24.35 \\
\midrule
\multicolumn{5}{c}{\textit{Overall Social Interaction: Macro F1}} \\
\cmidrule(lr){1-5}
Phi-4           & 51.65 & 67.28 & 56.32 & 72.08 \\
Gemini 1.5 Pro  & 53.16 & 54.71 & 53.50 & 58.64 \\
Gemini 2 Flash  & 56.08 & 63.30 & 56.41 & 67.83 \\
Gemini 2.5 Pro  & 53.97 & 61.15 & 54.63 &  61.46 \\
\bottomrule
\end{tabular}
\vspace{1mm}
\caption{Ablation studies on EgoSoD's components. APG: Audio Processing Graph, VPG: Video Processing Graph. Both APG and VPG positively contribute  to  the  overall  performance. And audio has a stronger contribution.}
\label{tab:comp} 
\vspace{-7mm}
\end{table}



\subsection{Different Graph Design Format}
 
 We conducted ablation studies on the EgoSocial dataset to evaluate different input formats of the graph. These included: Auto, which uses the raw cue questions; Graph, which provides predicted cue answers in a triplet graph format; -Dep, which prompts the model to rely heavily on the cues; -Think, which asks the model to provide the graph-of-thought and identify the cues used; and -H, which prompts for hierarchical thinking. As shown in Table~\ref{tab:graph_all}, our experiments indicate that hierarchical thinking is more effective for larger or more advanced models, likely due to their enhanced reasoning capabilities. And -Dep effectively helps the samll model to gain a remarkable improvement.
\begin{table}[!h]
\centering
\small
\begin{tabular}{lcccc}
\toprule
\textbf{Methods / Models} & \textbf{Phi4} & \textbf{Gemini 1.5 Pro} & \textbf{Gemini 2 Flash} & \textbf{Gemini 2.5 Pro} \\
\midrule
\multicolumn{5}{c}{\textit{Intervention Timing}} \\
\cmidrule(lr){1-5}
Baseline & 12.50 & 13.36 & 17.67 & 14.44 \\
Auto & 53.66 & 17.89 & 34.48 & 25.22 \\
Auto-H & 52.16 & 23.92 & 34.48 & 31.90 \\
Auto-Think & 53.88 & 18.10 & 35.78 & 26.29 \\
Auto-Think-H & 52.59 & 23.06 & 35.34 & 31.90 \\
Graph-Dep & 58.41 & 20.69 & 38.79 & 24.35 \\
Graph-Dep-H & 57.76 & 21.34 & 40.73 & 24.78 \\
Graph & 56.68 & 19.83 & 35.99 & 22.84 \\
Graph-H & 56.68 & 21.34 & 35.78 & 23.92 \\
Graph-Think & 55.60 & 20.91 & 36.85 & 23.06 \\
Graph-Think-H & 56.25 & 22.41 & 37.93 & 23.92 \\
\midrule
\multicolumn{5}{c}{\textit{Overall Social Interaction: Macro F1}} \\
\cmidrule(lr){1-5}
Baseline & 51.65 & 53.16 & 56.08 & 53.97 \\
Auto & 69.52 & 56.25 & 66.07 & 61.31 \\
Auto-H & 68.58 & 60.46 & 66.37 & 65.60 \\
Auto-Think & 69.81 & 56.47 & 66.49 & 62.06 \\
Auto-Think-H & 69.44 & 59.79 & 67.14 & 65.30 \\
Graph-Dep & 72.08 & 58.64 & 67.83 & 60.88 \\
Graph-Dep-H & 68.88 & 59.02 & 68.50 & 61.46 \\
Graph & 69.29 & 57.94 & 67.04 & 60.12 \\
Graph-H & 68.82 & 59.02 & 66.55 & 60.79 \\
Graph-Think & 69.34 & 58.59 & 66.71 & 60.28 \\
Graph-Think-H & 68.45 & 59.70 & 67.64 & 61.01 \\
\bottomrule
\end{tabular}
\vspace{1mm}
\caption{Ablation studies on Graph format on EgoSocial. Hierarchical thinking is more effective for larger or more advanced models, likely due to their enhanced reasoning capabilities. And -Dep effectively helps the samll model to gain a remarkable improvement.
Auto: offer the raw cue questions; Graph: offer the predicted cues answers in triplet format as a graph; -Dep: highly depends on the cues; -Think: ask to offer the thinking chain and which cues are used; -H: ask to hierarchical thinking. }
\label{tab:graph_all}
\end{table}

\subsection{Visualization}

\begin{figure}[htb]
    \centering
    \includegraphics[width=1.\linewidth]{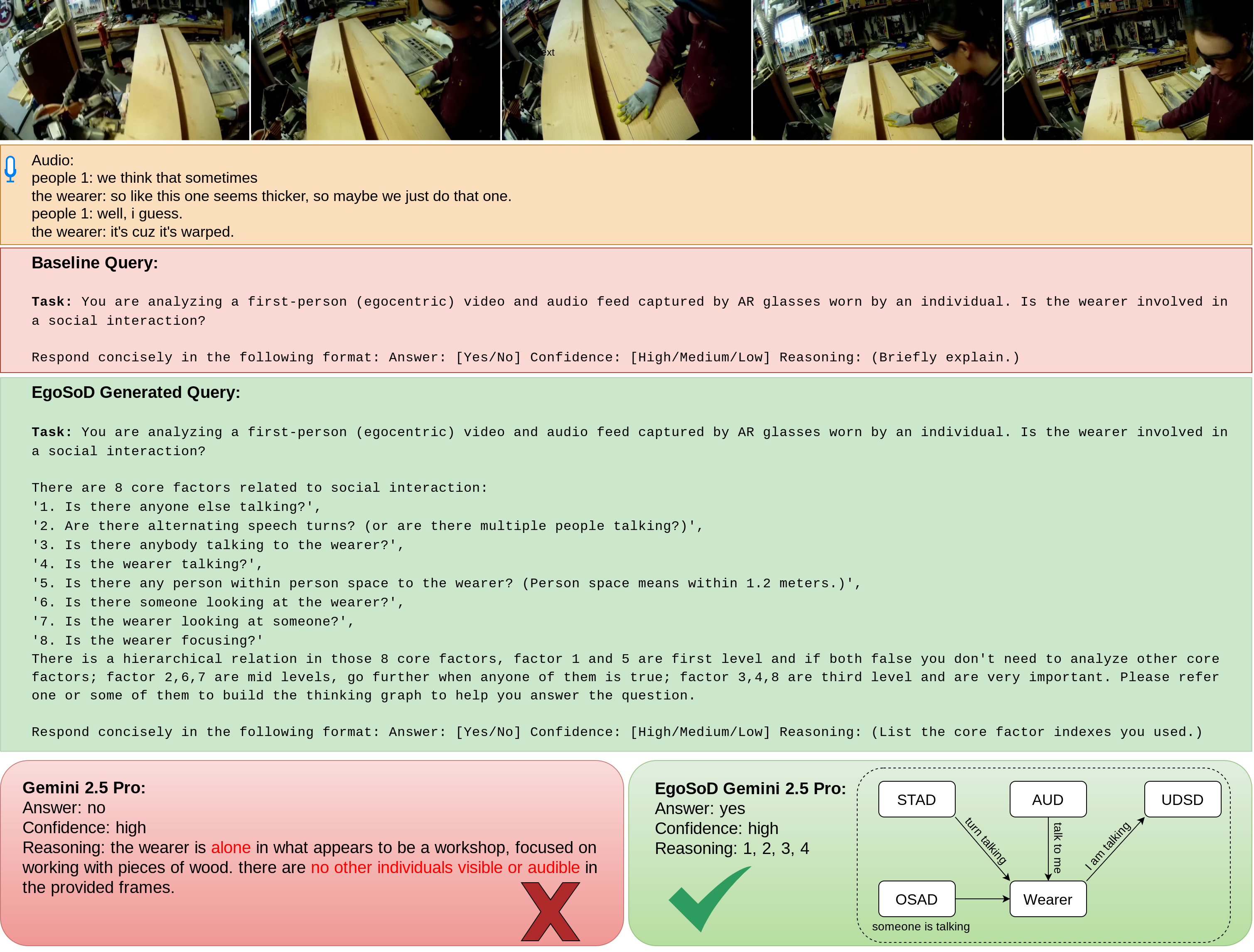}
    \vspace{-3mm}
    \caption{Demo of how EgoSoD helps OLLMs to detect social interaction. In this case, there are not that much valid visual information as there is neither looking at the wearer nor the wearer is looking at someone. So our EgoSoD guide the model to focus more on the audio part, which effectively outputs the right answer.}
    \label{fig:demo}
\end{figure}

As shown in Figure~\ref{fig:demo}, EgoSoD guides OLLMs in social interaction detection, especially when visual information is limited. In this example, neither the wearer nor anyone else is engaged in eye contact, reducing visual cues. EgoSoD compensates by emphasizing the audio input, which enables the model to accurately determine the social context.

\subsection{Different Amount of Vision Information}
Ablation studies on how much visual information needed. In our experiments, we offer 3 frames, 6 frames, and 10 frames. As shown in Table~\ref{tab:frame_num}, the visual information won’t dramatically affect the performance, while fewer visual information even better in some cases. This may because the current models are still struggling on handling large amount of visual information.
\begin{table}[htbp]
\centering
\small
\begin{tabular}{lcccc}
\toprule
\textbf{Frame-num / Models} & \textbf{Phi4} & \textbf{Gemini 1.5 Pro} & \textbf{Gemini 2 Flash} & \textbf{Gemini 2.5 Pro} \\
\midrule
\multicolumn{5}{c}{\textit{Intervention Timing}} \\
\cmidrule(lr){1-5}
3 frames & 58.63 & 22.63 & 37.50 & 26.52 \\
6 frames & 57.54 & 20.26 & 39.87 & 27.16 \\
10 frames & 57.76 & 21.34 & 40.73 &  24.78 \\
\midrule
\multicolumn{5}{c}{\textit{Overall Social Interaction: Macro F1}} \\
\cmidrule(lr){1-5}
3 frames & 69.50 & 59.80 & 66.84 & 62.38 \\
6 frames & 68.97 & 58.16 & 68.26 & 62.88 \\
10 frames & 68.88 & 59.02 & 68.50 & 61.46 \\
\bottomrule
\end{tabular}
\vspace{1mm}
\caption{Ablation studies on how much visual information needed.  The visual information won’t dramatically affect the performance, while fewer visual information even better in some cases. This may because the current models are still struggling on handling large amount of visual information.}
\label{tab:frame_num}
\end{table}

\section{Annotation Details}

This section provides a comprehensive overview of our annotation process, which was crucial for establishing ground truth and identifying various social cues relevant to our study. We combined two existing Ego4D annotations with five newly manually annotated cues, resulting in eight social cues per video segment. Throughout this section, "me" refers to the person wearing the glasses in the video.

\subsection{Existing Annotations from Ego4D}

The Ego4D dataset already includes the following questions:
\begin{itemize}
    \item \textit{Addressed-to-User Detection (AUD): Is someone talking to me?}
    \item \textit{Incoming Gaze Detection (IGD): Is someone looking at me?}
\end{itemize}

\subsection{Manual/Derived Annotations}

We meticulously defined and manually annotated six additional cues to enrich our dataset:
\begin{itemize}
    \item \textbf{\textit{Other-Speaker Activity Detection (OSAD): Is someone else talking?}} This was defined as any frame where anyone other than wearer is talking.
    \item \textbf{\textit{Speech-Turn Alternation Detection (STAD): Are people talking in turns?}} This was defined as true within a segment if at least two individuals spoke in turns.
    \item \textbf{\textit{User-Directed Speech Detection (UDSD): Am I talking?}} This was defined as any frame where the wearer was talking.
    \item \textbf{\textit{Proxemic Awareness Detection (PAD): Are people in personal space?}} Based on prior research \cite{barua2021detecting} highlighting personal space as a strong indicator of social interaction, we manually annotated this cue as true when individuals were in a group within an approximate distance of 1.2 meters or less.
    \item \textbf{\textit{Outgoing Gaze Detection (OGD): Am I looking at someone?}} This was annotated if the wearer was gazing at another person.
    \item \textbf{\textit{Self-Focus Detection (SFD): Am I focusing on something?}} This was defined as any frame where the wearer was concentrating on an object or activity—for instance, shuffling cards or contemplating a move in a game.
    
\end{itemize}
To ensure the high quality of all manual labels, we employed three independent annotators. We then applied a majority voting scheme, accepting annotations where at least two annotators agreed with high confidence and discarding ambiguous videos.

\subsection{Ground Truth Definition}

For our primary objective, we define social interaction as an active conversation where a user would least likely want to be disturbed. We consider an active conversation to be taking place when the wearer is speaking to someone, or someone is speaking to the wearer. This is determined by the presence of either \textit{Is someone talking to me?} or \textit{Am I talking?} cues within a 10-second segment. All other scenarios, including challenging ones like being in a group without direct interaction, are designated as negative samples.

Refer to Figure~\ref{fig:data_distribution} for the complete distribution of all social cues and the ground truth established for our task.

\subsection{Dataset Construction and Processing}

Our dataset was constructed using videos from the Ego4D dataset, which captures diverse real-world egocentric multi-user scenarios, including games, eating/drinking, and more. We selected a specific subset from this larger collection to create our evaluation dataset.

Each original Ego4D clip is 5 minutes long and comes with detailed annotations. To provide optimal context for OLLM models to make real-time decisions and facilitate standardized analysis, we segmented these clips into 10-second chunks. This duration offers sufficient context for models to make informed decisions without becoming computationally overwhelming. We downsampled the video frequency to 1 frame per second (1fps) to align with real-time processing requirements. We also ensured that the audio and existing Ego4D frame-level annotations were correspondingly divided and aligned with these 10-second video segments. This processing resulted in a refined dataset comprising 1500 video segments. With these 1500 processed video segments, our annotation process generated a comprehensive set of 13,500 video-question pairs. Our goal in curating these specific cues is to rigorously analyze the underlying effectiveness of current OLLM models in understanding complex social dynamics.

\begin{figure}[h]
    \centering
    \includegraphics[width=1.\linewidth]{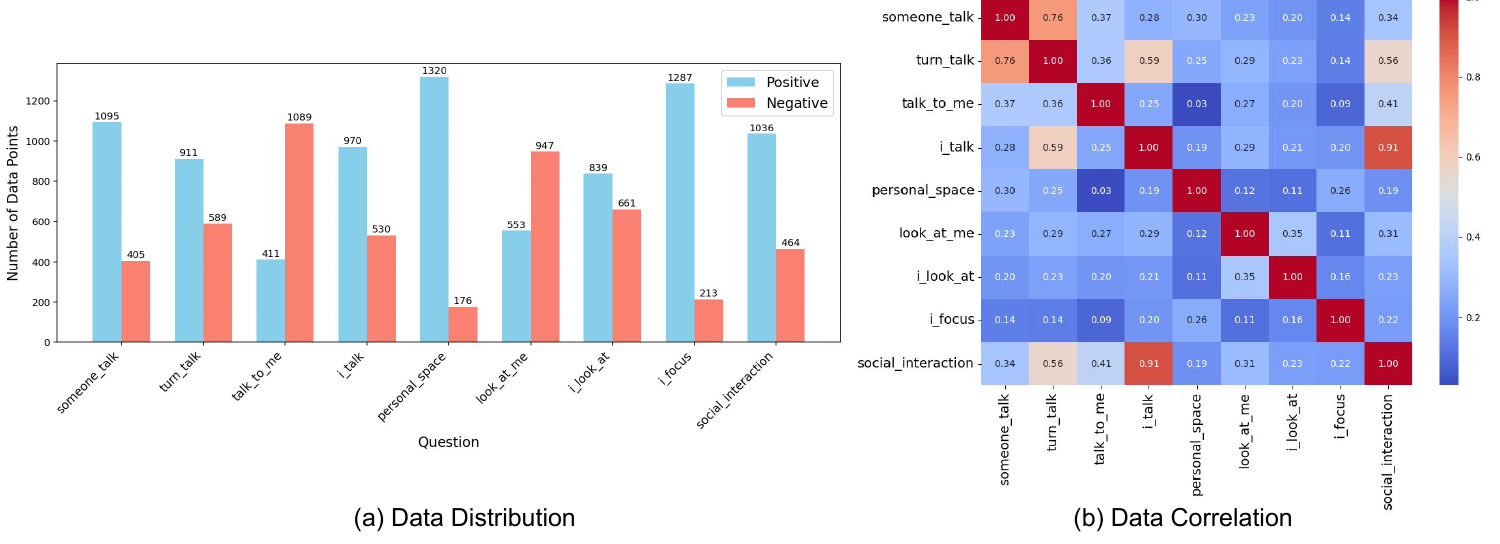}
    \caption{Data distribution and correlation on our EgoSocial dataset.}
    \label{fig:data_distribution}
    \vspace{-4mm}
\end{figure}

\subsection{More Discussion}
To enable social interaction detection to adapt its behavior appropriately across diverse environments, Scenario Analysis is crucial. This helps the model analyze contextual cues to identify the nature of the interaction setting.

For augmented reality (AR) or virtual reality (VR) environments, the model will consider the presence of virtual avatars, spatial constraints, and the integration of digital elements with the real world. In gaming contexts, the model will recognize game-specific objectives, rule sets, and player roles. For virtual meetings, the analysis will focus on cues such as structured agendas, presentation sharing, and turn-taking protocols.

In a conference scenario, the model's sensitivity should be higher since users typically need to focus more. Conversely, in a cafe scenario, the model's sensitivity can be lower due to the relaxed social environment. By discerning these distinct characteristics, the model can tailor its interaction strategies, including non-verbal communication, conversational styles, and collaborative behaviors, to align with the specific demands of each scenario.

{\small
\bibliographystyle{ieee_fullname}
\bibliography{mainbib}
}

\end{document}